\documentclass{article}

\usepackage{arxiv}

\usepackage[utf8]{inputenc} 
\usepackage[T1]{fontenc}    
\usepackage{hyperref}       
\usepackage{url}            
\usepackage{booktabs}       
\usepackage{amsfonts}       
\usepackage{nicefrac}       
\usepackage{microtype}      
\usepackage{lipsum}
\usepackage{graphicx}
\usepackage{caption}
\usepackage{subcaption}
\usepackage{multirow}
\usepackage{algorithm}  
\usepackage{algpseudocode}  

\title{Design of an Optoelectronically Innervated Gripper for Rigid-Soft Interactive Grasping\thanks{This work was supported by Southern University of Science and Technology and AncoraSpring Inc. }}

\author{
  Linhan Yang$^{\#}$\\
  Department of Mechanical and Energy Engineering\\
  Southern University of Science and Technology\\
  Shenzhen, China 518055\\
  \texttt{yanglh9652@gmail.com}\\
  \And
  Xudong Han$^{\#}$\\
  Department of Mechanical and Energy Engineering\\
  Southern University of Science and Technology\\
  Shenzhen, China 518055\\
  \texttt{11812519@mail.sustech.edu.cn}\\
  \And
  Weijie Guo\\
  Department of Mechanical and Energy Engineering\\
  Southern University of Science and Technology\\
  Shenzhen, China 518055\\
  \texttt{11510615@mail.sustech.edu.cn}\\
  \And
  Zixin Zhang\\
  Department of Mechanical and Energy Engineering\\
  Southern University of Science and Technology\\
  Shenzhen, China 518055\\
  \texttt{11711101@mail.sustech.edu.cn}\\
  \And
  Fang Wan\\
  Department of Mechanical and Energy Engineering\\
  Southern University of Science and Technology\\
  Shenzhen, China 518055\\
  \texttt{sophie.fwan@hotmail.com}\\
  \And
  Jia Pan\\
  Department of Computer Science\\
  University of Hong Kong\\
  Hong Kong, China 999077\\
  \texttt{jpan@cs.hku.hk}\\
  \And
  Chaoyang Song\thanks{Corresponding Author.}\\
  Department of Mechanical and Energy Engineering\\
  Southern University of Science and Technology\\
  Shenzhen, China 518055\\
  \texttt{songcy@ieee.org}\\
}

\begin{document}
\maketitle

\begin{abstract}
    
    Over the past few decades, efforts have been made towards robust robotic grasping, and therefore dexterous manipulation. The soft gripper has shown their potential in robust grasping due to their inherent properties-low, control complexity, and high adaptability. However, the deformation of the soft gripper when interacting with objects bring inaccuracy of grasped objects, which causes instability for robust grasping and further manipulation. In this paper, we present an omni-directional adaptive soft finger that can sense deformation based on embedded optical fibers and the application of machine learning methods to interpret transmitted light intensities. Furthermore, to use tactile information provided by a soft finger, we design a low-cost and multi degrees of freedom gripper to conform to the shape of objects actively and optimize grasping policy, which is called Rigid-Soft Interactive Grasping. Two main advantages of this grasping policy are provided: one is that a more robust grasping could be achieved through an active adaptation; the other is that the tactile information collected could be helpful for further manipulation.
    
\end{abstract}


\section{Introduction}
\label{sec:Introduction}
    
    Robust grasping and control contribute to manipulation. Recent research on robust grasping presents a growing interest in using computer vision for grasp prediction \cite{pinto2016supersizing, lenz2015deep, mahler2019learning}, high-resolution tactile sensor \cite{yuan2017gelsight, yamaguchi2016combining} and sophisticated design of gripper \cite{ma2017yale, yuan2020design}. 

    With the advances in computer visions, there has been a recent paradigm shift in robotic grasp planning to the data-driven learning method \cite{bohg2013data}. A great number of data set have been proposed with human-labeled similarly in computer vision field \cite{lenz2015deep}, \cite{chu2018real}, with model-based synthetic data set \cite{mahler2019learning,mahler2017dex}, or using physical trials directly \cite{pinto2016supersizing,levine2018learning}. However, when humans manipulate objects, visual information is not the only stimuli they rely on. Tactile information plays an even more important role when humans are executing manipulation skills, especially when visual information is not precise or partly blocked \cite{fazeli2019see}. There has been evidence that the robot can grasp without seeing \cite{murali2018learning}. Some vision-based sensors have been proposed to acquire tactile information \cite{yuan2017gelsight, yamaguchi2016combining}. They take pictures of the sensor surface using a camera and transfer this visual data to tactile data such as deformation and force. Proprioception is another way to get plentiful tactile information to interpret end effector deformation to the tactile sensor. Soft finger with compliant materials is a feasible choice for end effector with proprioception \cite{van2018soft, thuruthel2019soft}.

    In this paper, we focus on interpreting tactile information from our soft finger when interacting and optimize grasping policy after an initial grasp based on tactile information.

\begin{figure}[t]
    \begin{centering}
    \textsf{\includegraphics[width=0.5\columnwidth]{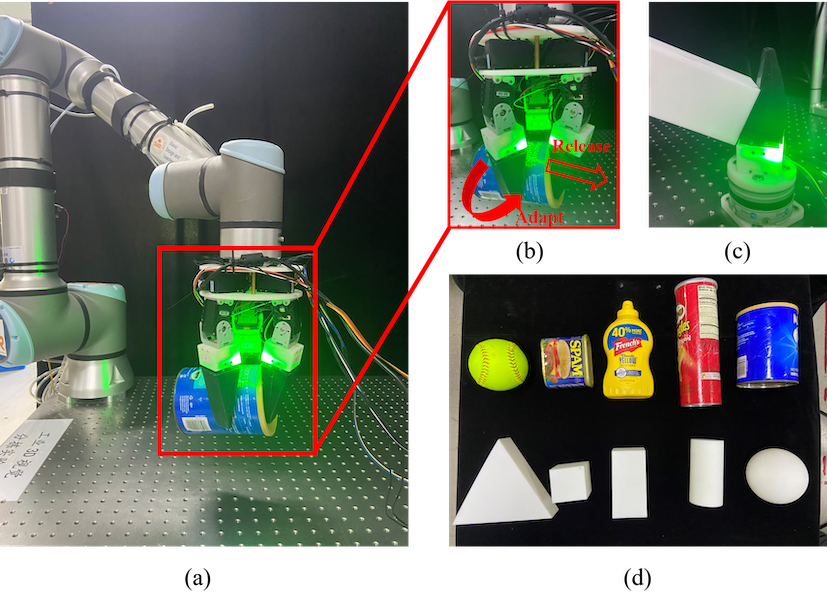} }
    \par\end{centering}
    \caption{Overall of this paper: (a) Initial grasp with  optoelectronical  innervated  gripper; (b) Adjustment after initial grasp; (c) Calibration of optical fiber-based soft finger (d) Objects for calibration and testing}
    \label{fig:Overall}
\end{figure}

\subsection{Related Work}
\subsubsection{Soft Robotic Grippers}

    The soft gripper has shown potential in grasping for its high adaptability and low control complexity \cite{shintake2018soft}. Compared to the rigid ones, the soft finger tends to deform to adapt to objects' shapes due to the compliant materials. The deformation process is passive \cite{yang2020rigid, wan2020reconfigurable} or under-actuated \cite{ma2017yale, odhner2014compliant, deimel2016novel} due to the material softness and mechanical compliance such that control complexity is mostly reduced. Most of these fingers are actuated by only one actuator or pneumatic, and the form closure of the grasped object is easily achieved. Precision grasping is also investigated. Dollar et al. \cite{ma2017yale, odhner2014compliant} proposed a series of the underactuated gripper, which could transform between power-grasping, spherical-grasping, and lateral-grasping models. However, all of the control algorithms between different grasping models are open-loop and hardcoded, not applied in an unstructured environment.

\subsubsection{Proprioceptive Sensors}

    Stretchable sensors such as soft resistive sensors, soft optoelectronic sensory have been embedded in soft fingers to sense the deformation of soft fingers when interacting with the environment \cite{van2018soft, thuruthel2019soft, zhao2016optoelectronically}. Zhao et al. \cite{zhao2016optoelectronically} proposed a prosthetic hand using optical waveguides for strain sensing, which could be used to sense the shape and softness of selected objects. Meerbeek et al. \cite{van2018soft} present an elastomeric foam with 30 optical fibers internally illuminated. They trained this foam sensor system with a machine learning method to sense when the foam is bent and twisted. Thuruthel et al. \cite{thuruthel2019soft} proposed a human-inspired soft pneumatic actuator with embedded soft resistive sensors. They model the kinematics of this soft continuum actuator in real-time with recurrent neural networks. These research show the great potential of the soft compliant gripper with proprioception. However, all of these research stay in the fundamental stage and are not applied to real grasping or manipulation scenario. 

\begin{figure}[t]
    \begin{centering}
    \textsf{\includegraphics[width=0.5\columnwidth]{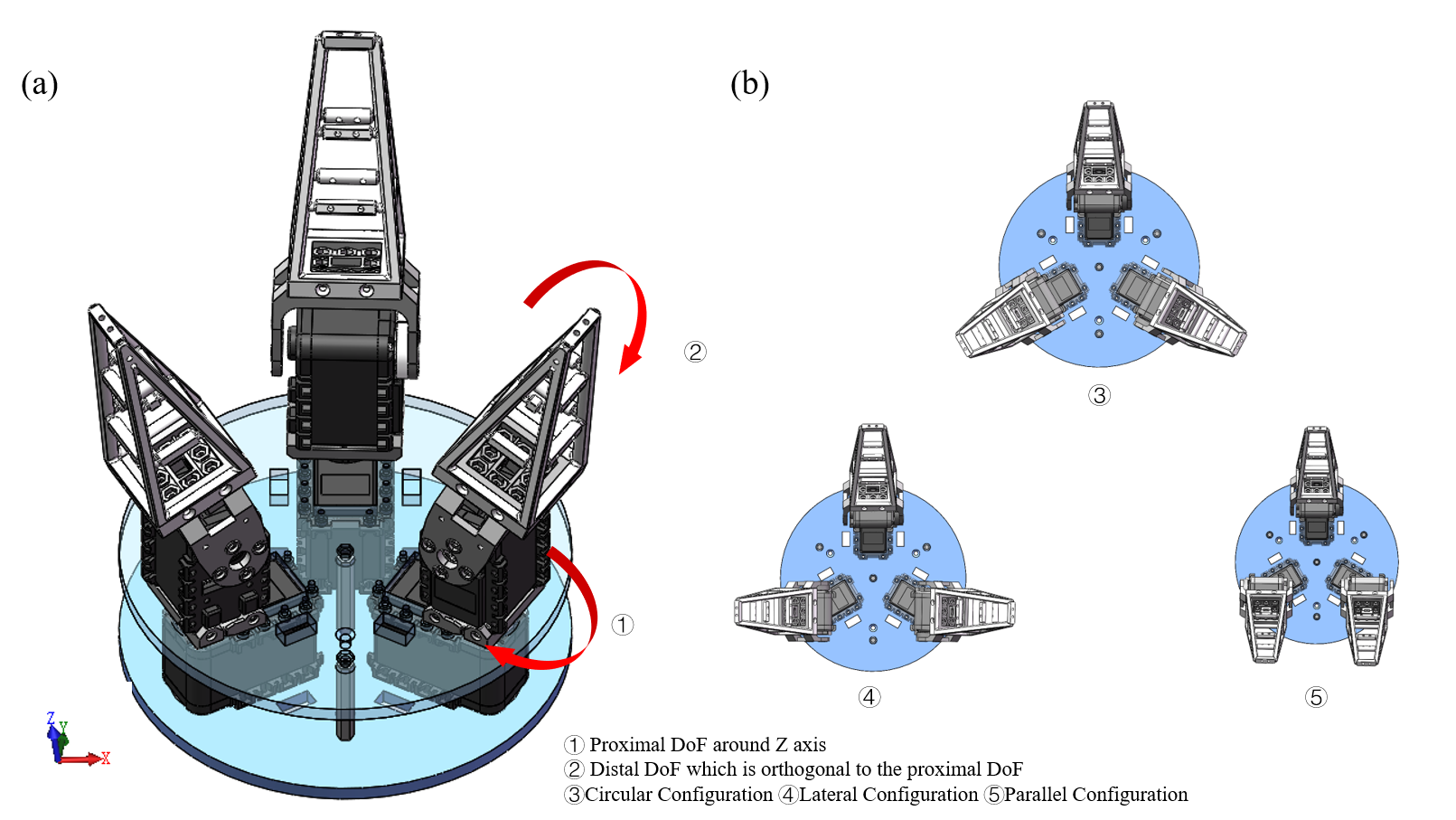}}
    \par\end{centering}
    \caption{CAD model of gripper: (a) three fingers each with two degrees of freedom have (b) three different configurations for grasping.}
    \label{fig:CADModel}
\end{figure}

\begin{figure}[t]
    \begin{centering}
    \textsf{\includegraphics[width=0.5\columnwidth]{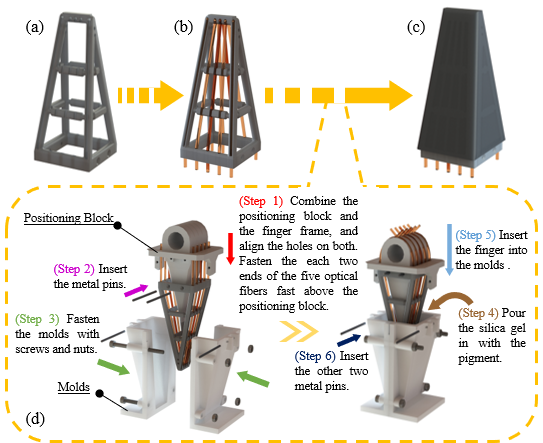}}
    \par\end{centering}
    \caption {The design and fabrication process of the optical fiber-based soft finger. (a) shows finger frame. (b) shows finger frame with the fibers. (c) shows finger frame with silica gel  skin. (d) shows the details of the making process of black silica gel skin. (In order to express the transparent fibers clearly, the orange transparent material is used in these figures instead.)}
    \label{fig:Finger}
\end{figure}

\subsection{Proposed Method and Contributions}

    First, we present a novel soft finger structure, which has an omni-directional adaptation to the environment. Such a finger is capable of grasping a wider variety of objects which different shapes and sizes. Five optical fibers are embedded in the interior of the finger to sense the deformation when interacting. Initial grasping is open-loop and used for collecting the tactile information, including the shape and size of target objects. Then, in order to make use of tactile information for further grasping adjustment, we design a multi-grasping-model modular gripper, which will be introduced in section \ref{sec:Method}. This gripper has six degrees of freedom and is capable of both precise grasping and power grasping. Based on the tactile information, which is normal force and torque in this paper, we optimize grasping mode by minimizing the lateral friction and torque, increasing grasping quality. This closed-loop process is called rigid-soft interactive grasping.

    This paper's contribution is to demonstrate an optical fiber-based soft finger with proprioception and a modular gripper consisting of five former fingers that could grasp the target objects and adapt the shape based on tactile information provided by soft fingers. 

    In the rest of this paper, Section \ref{sec:Method} explains the proposed method and problem formulation of rigid-soft interactive grasping. Experimental results are enclosed in section \ref{sec:ExpResults}, which is followed by section \ref{sec:Discussion} and \ref{sec:Conclusion} that ends the paper.

\section{Method}
\label{sec:Method}
\subsection{Gripper Design and Assembly}
    
    The gripper consists of three identical fingers, and each has two actuated DoFs. ROBOTIS Dynamixel smart actuators provide all of these six DoFs. In order to obtain excellent reliability and reduce the cost, we choose ROBOTIS Dynamixel AX-12A as our actuators, which features fully integrated motors with low price and small size. For each finger, two actuators are daisy-chained, and each can provide a rotational joint around the same or different axis. A 12V power supply is used to power the gripper, which helps the actuators output enough torque to complete the grasping action. High-level Python API provided by Dynamixel SDK is used for communication between the gripper hardware and host computer. The information transferred includes the current position, goal position, and torque limit for adjusting the finger's stiffness when interacting with different objects. The whole gripper only consists of off-the-shelf components (actuators, printed circuit boards, and optical fibers) and 3D printing products (soft fingers and other assembly parts), which cost several hours replicate.

    The CAD model of the gripper is shown in Fig. \ref{fig:CADModel}. Each daisy-chained actuator provides proximal and distal DoFs. The proximal DoF is achieved by a rotational joint around Z-axis, while the distal DoF is orthogonal to the proximal DoF. For demonstration, we illustrate this gripper with three grasping configurations: Circular Configuration, where three fingers are arranged circularly, which facilitating the grasp of a spherical object; Lateral Configuration, where two fingers are arranged symmetrically, and the third one is placed such that the proximal joint axis is on the mid-plane of the others, which facilitate the manipulation and precise grasp; Parallel Configuration, where two fingers are on the same side and parallel to the third finger, which facilitating the grasp of a cylindrical object. Such a gripper can adapt any surface of target objects using the three typical configurations.

\subsection{Optical Fiber-based Soft Finger Design and Fabrication}

    A novel soft finger structure, which is inspired by \cite{yang2020rigid, wan2020reconfigurable}, is used to be our gripper's end-effector, as shown in Fig. \ref{fig:Finger}. Such a structure's advantage is its inherent omnidirectional adaptation to any shapes of objects, ultra-low-cost, and high environmental suitability. The deformation of such a soft finger occurring when interacting with the environment can essentially provide tactile information with an infinite degree of freedom. Such deformation can be thought of as environmental perception, which in turn guides interaction with the environment.

    To interpret the deformation, we demonstrate an optical fiber array, which consists of five optical fibers to estimate the curvature, as shown in Fig. \ref{fig:Finger}(a)(b). The luminous flux loss correlates with the curvature of optical fiber, which makes the optical fibers a feasible choice to interpret soft fingers' deformation. Five LEDs are placed at the optical fibers' start as the luminous transmitter, and five photo resistance is placed at the receiving end to estimate the output luminous intensity. To reduce the ambient light's impact to get enough luminous flux, we choose 520-525NM LEDs and 520-550NM photo resistance, whose wavelength segments are more concentrated and match each other, which significantly increase the signal-to-noise ratio for the required signals. Moreover, after the optical fiber array placed on the finger framework, the surface is covered with black silica gel skin for isolation from the environment, as shown in Fig. \ref{fig:Finger}(c), which is also to increase the signal-to-noise ratio.

    The skin is made by Dragon Skin 10 MEDIUM silica gel, whose strength is suitable. However, this silica gel is originally a milky white translucent liquid, which means it needs to be mixed with black pigment at a ratio of 20:1 to isolate the ambient light effectively. Moreover, the silica gel skin is set 3mm away from the finger's outer surface to increase the finger surface texture to enhance the grasping effect for future work. Fig. \ref{fig:Finger} (d) shows the details of the making process of black silica gel skin.

    Let $I_{0}$ denote the baseline luminous intensity without any deformation. With the current output luminous intensity $I$, the luminous flux loss in decibels through the optical fiber is then described as

\begin{equation}
    \label{eq:luminous flux loss}
     a = 10 log_{10}(I_{0}/I)
\end{equation}

    By this definition, the output loss $a$ is 0 without deformation and less than 0 when interacting with the environment.As for our finger, $a=(a_{1}, a_{2}, a_{3}, a_{4}, a_{5})\in R^{5}$ is denoted to be the luminous flux loss of all five optical fibers. This vector $a$ can typically describe the deformation of the finger because of the positive correlation between the luminous flux loss and the curvature.

    It is necessary to establish a deformation-and-reaction map. Therefore, during the calibration experiment, while receiving the finger deformation data, we also need to obtain the reaction caused by interacting with the environment. OnRobot HEX-E v2 is used to capture force and torque data in X-, Y-, and Z-axes. The finger's bottom is attached to OnRobot HEX-E v2, which establishes 6D reaction data referenced to the coordinate system whose origin is the bottom center of the finger. OnRobot HEX-E v2 provides a vector, $r = (F_{x}, F_{y}, F_{z}, T_{x}, T_{y}, T_{z})$, containing force and torque data in the X-, Y-, and Z-axes.

    When the finger interacts with the environment, deformation and force data are collected simultaneously. If combining the vector composed of the luminous flux loss data obtained from the photoresistance group with the vector composed of the six-dimensional force data obtained from OnRobot HEX-E v2, we can get the following vector.

$ v_{sensor} = (a_{1}, a_{2}, a_{3}, a_{4}, a_{5}, F_{x}, F_{y}, F_{z}, T_{x}, T_{y}, T_{z}) $

    We can get many vectors in the above form by taking several different objects to interact with the finger, which is used as training sets for machine learning to obtain a deformation-and-reaction map.

\begin{figure}[t]
    \begin{centering}
    \textsf{\includegraphics[width=0.5\columnwidth]{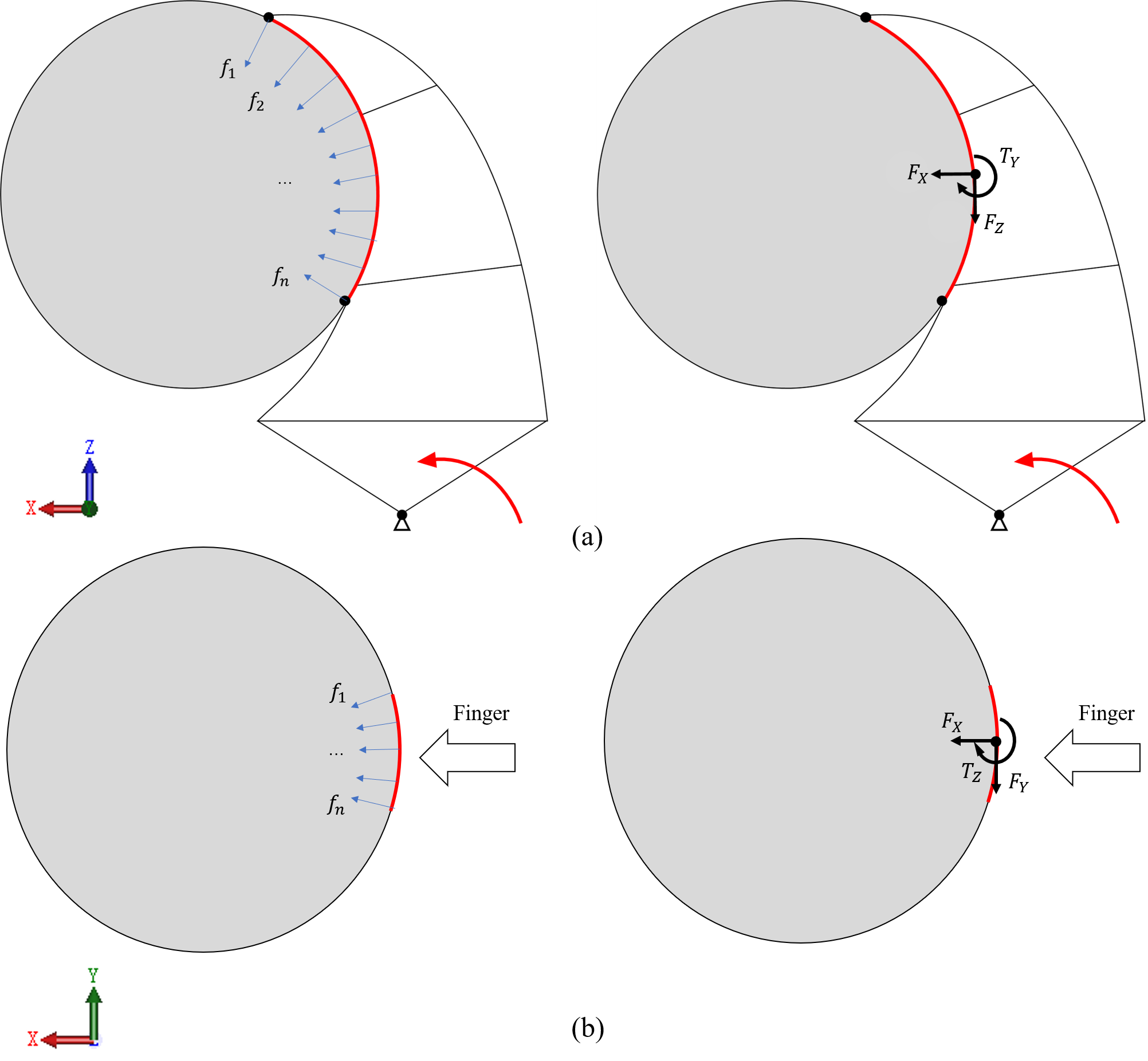} }
    \par\end{centering}
    \caption{Schematic of single finger contact in (a) X-Z cross section and (b) X-Y cross section.}
    \label{fig:SingleFinger}
\end{figure}

\begin{figure}[htbp]
    \begin{centering}
    \textsf{\includegraphics[width=0.5\columnwidth]{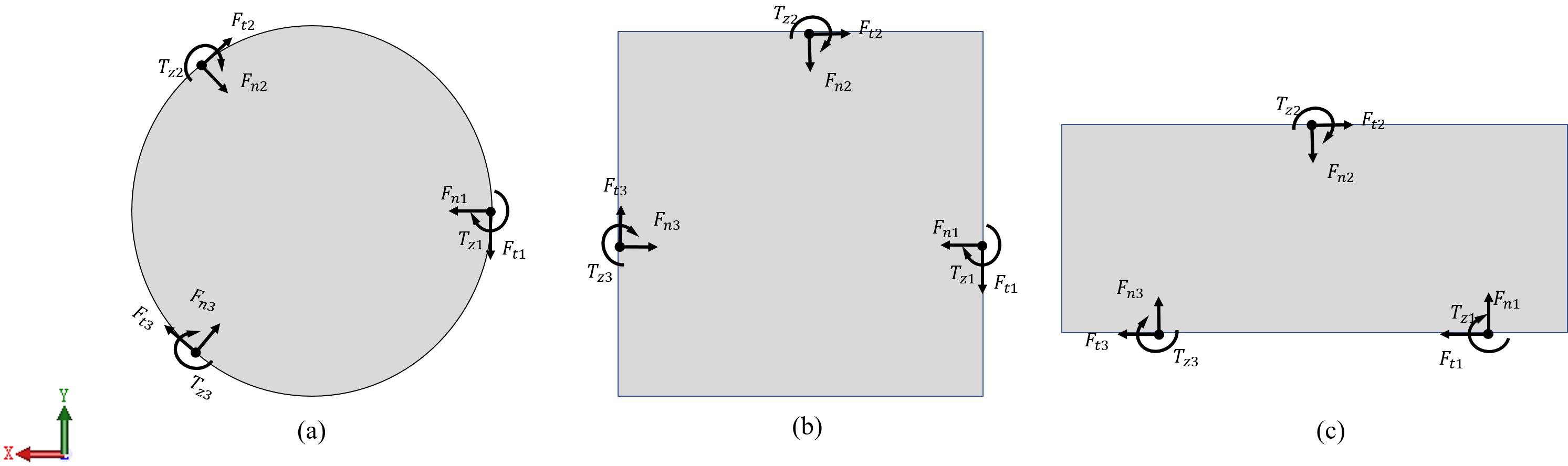} }
    \par\end{centering}
    \caption{Schematic of grasp model: (a) Circular Configuration; (b) Lateral Configuration; (c) Parallel Configuration.}
    \label{fig:GraspModel}
\end{figure}

\begin{algorithm}[t] 
\caption{Rigid-Soft Interactive Grasping.}
 \label{alg:Rigid-Soft}  
  \begin{algorithmic}[1]  
     \Require  
    Base Configuration $\pi$, Normal force $F_{n,i}$, Torque $T_{z,i}, i=1,2,3$.
     \Ensure  
    Optimized Grasp.
    
    \State  Obtain Normal force $F_{n,i}$, Torque $T_{z,i}$ $ \gets Initial Grasp$
    \State  \Return $F_{n,i}$, Torque $T_{z,i}$.
     
    \Function{Grasp Optimization}{$F_{n,i}, T_{z,i}$}
        \While{$T_{z,i} \neq 0$, for any $i =1,2,3 $}
             \State Torque Optimization($T_{z,i}$). Objective: $T_{z,i} = 0$ , for all $i =1,2,3$
             \State \Return Base Configuration $\pi$
        \EndWhile
        \While{$\stackrel[i]{}{\sum}F_{n_{0},i} \neq 0, i=1,2,3 $} 
             \State Friction optimization ($F_{n,i}$, Base Configuration $\pi$), Objective: $\stackrel[i]{}{\sum}F_{n_{0},i} = 0, i=1,2,3 $
             \State \Return Optimized Normal force $F_{n,i}$
        \EndWhile
    \EndFunction
  \end{algorithmic}
\end{algorithm}

\subsection{Problem Formulation}
    
    The goal of this problem is to optimize grasping quality after initial grasping based on the deformation of gripper. 

    The contact between gripper and object when interacting is obviously planar. We simplify this problem by discretizing contact planar as several or single point for computational convenience. Next, single finger contact model and multi-finger grasping model are discussed respectively. 
    
\subsubsection{Single finger contact model}
    
    We assume the following information:
\begin{enumerate}
    \item The deformation of finger when interacting is large enough to conform the surface of object such that normal vector of contact force is orthogonal to surface of object. 
    \item Torque along the X- and Y-axis when interacting is neglected for its small magnitude and little influence to final grasping performance. 
\end{enumerate}

    In the simplified single finger case illustrated in Fig. \ref{fig:SingleFinger}, the contact force is discretized as $F = (f_{1}, f_{2}, ..., f_{n})$. In the X-Z cross section shown in Fig. \ref{fig:SingleFinger} (a), the resultant force of contact is capable of transmitting a normal force $F_{x}>0$, a frictional force $F_{z}$ and balancing torque $T_{y}$ along Y-axis, which is neglected in assumption. In the X-Y cross section shown in Fig. \ref{fig:SingleFinger} (b), the resultant force of contact is capable of transmitting a normal force $F_{x}>0$, a frictional force $F_{y}$ and balancing torque $T_{z}$ along Z-axis.

    According to Coulomb friction model, when object grasped: 
\begin{equation}    \label{eq:Coulomb friction}
	\left\| F_{t} \right\| \leq \mu F_{n}, v = 0
\end{equation}
    where $\mu$ is friction coefficient and $v$ is the sliding velocity of object. If the sliding velocity is zero, then the magnitude of the tangential friction force is less than or equal to $\mu$ times the normal force, which is non-negative. As for the schematic shown in Fig. \ref{fig:SingleFinger}, the normal force $F_{n} = F_{x}$ and the tangential frictional force $F_{t} = F_{y}+ F_{z}$. Then we get:

\begin{equation}    \label{eq:Coulomb friction2}
	\left\| F_{y}+ F_{z} \right\| \leq \mu F_{x}, v = 0
\end{equation}

\subsubsection{Grasp model}

    In this paper, we have three contact area, one for each finger, illustrated in Fig. \ref{fig:GraspModel}. According to Nowtonian mechanics equilibrium equation, when sliding velocity $v = 0$, we get:

\begin{equation}    \label{eq:Nowtonian mechanics1}
	\stackrel[i]{}{\sum}F_{n,i} +\stackrel[i]{}{\sum}F_{t,i} + F_{ext} = 0,
\end{equation}

    and

\begin{equation}    \label{eq:Nowtonian mechanics2}
	\stackrel[i]{}{\sum}T_{z,i} + T_{ext} = 0, i = 1,2,3
\end{equation}

    Derived from Eq. \ref{eq:Coulomb friction}, we get:

\begin{equation}    \label{eq:Nowtonian mechanics3}
	\left\| F_{t,i} \right\| \leq \mu F_{n,i}, i = 1,2,3
\end{equation}

\subsection{Rigid-Soft Interactive grasping}

    The objective of our optimization policy is to get a robust grasp, that is to maintain the equilibrium equation when there is an external disturbance. Given a constant $F_{n,i}, i=1,2,3$, when there is no external disturbance, 

\begin{equation}    \label{eq:equilibrium equation with no disturbance}
    \stackrel[i]{}{\sum}F_{t_{0},i} = -\stackrel[i]{}{\sum}F_{n_{0},i}
\end{equation}

    Then, the anti-disturbance ability could be defined as minimum of $F_{ext}$ which could break the equilibrium state:
\begin{equation}    \label{eq:min Fext equation}
    F_{ext} =  \max \stackrel[i]{}{\sum}F_{t,i} - \stackrel[i]{}{\sum}F_{t_{0},i} , i=1,2,3
\end{equation}
    where $\max F_{t,i} = \mu F_{n,i},i=1,2,3$. To maximize anti-disturbance ability, $ F_{t_{0},i} = 0,i=1,2,3$. This process is called friction force optimization. Then, we get the objective equation of friction force optimization:

\begin{equation}    \label{eq:objective equation of normal direction}
    \stackrel[i]{}{\sum}F_{n,i} = 0, i=1,2,3
\end{equation}

    Similarly, as for torques along Z-axis, the objective equation is:

\begin{equation}    \label{eq:objective equation along Z-axis}
    T_{z,i} = 0, i=1,2,3, 
\end{equation}
    where $T_{z}$ is essentially caused by twist of soft finger and smaller $T_{z}$ could bring a more stable system. This process is called torque optimization. Also, if $T_{z} = 0$, we can calculate direction of $F_{n_{0}}$ by base configuration of gripper.

    We summary this whole optimization process in Algorithm \ref{alg:Rigid-Soft}, called Rigid-Soft Interactive Grasping.

\section{Experiment Results}
\label{sec:ExpResults}

\subsection{Calibration}

    The prerequisites of our optimization algorithm is to get $F_{x}, T_{z}$ of each finger. To calibrate our finger, in other words, to establish a map $f: a = (a_{1}, a_{2}, a_{3}, a_{4}, a_{5}) \rightarrow \tilde{r}=(F_{n}, T_{z}) $, we push a set of standard objects, as shown in Fig.\ref{fig:Overall} (d) , against soft finger with random continuous external force magnitude in arbitrary direction to simulate real grasp. Force and torque data is collected by OnRobot HEX-E v2 mounted on the bottom of soft finger, as shown in Fig \ref{fig:Overall} (c).

    Through the machine learning method provided by \cite{scikit-learn}, we establish the deformation-and-reaction map. To measure the accuracy of the established model, we use two indicators, RMSE and R Squared, which are often used to measure linear regression quality. Root Mean Square Error (RMSE) is the standard deviation of the residuals, reflecting the model's accuracy. And the equation is 

\begin{equation}    \label{eq:RMSE}
    RMSE=\sqrt{\frac{1}{m}\sum_{i=1}^{m}\left(y^{\left(i\right)}-{\hat{y}}^{\left(i\right)}\right)^2}
\end{equation}

    R Squared (R$^2$) is another useful indicator of goodness of fit, ranging from 0 to 1. The closer the value is to 1, the better the quality of the model. The equation is

\begin{equation}    \label{eq:R2}
    R^2=1-\frac{\sum_{i=1}^{m}\left({\hat{y}}^{\left(i\right)}-y^{\left(i\right)}\right)^2}{\sum_{i=1}^{m}\left({\bar{y}}^{\left(i\right)}-y^{\left(i\right)}\right)^2}
\end{equation}

    Since in the grasping optimization policy, to support grasping adjustment, our goal is to make $T_z$ become 0, only the sign of $T_z$ needs to be considered instead of the magnitude. So classification is made for $T_z$: we just classify whether the sign of $T_z$ is positive, negative, or 0, in order to determine which direction the actuator needs to turn to make $T_z$ zero. The prediction performance of regression and classification for three fingers is shown in Table. \ref{tab:prediction performance table}, in which Root Mean Square Error (RMSE), whose expression is shown as Eq. \ref{eq:RMSE}, and R Squared (R$^2$), whose expression is shown as Eq. \ref{eq:R2}.

\begin{table}[t]
    \centering
    \caption{Prediction Performance of Regression and Classification for Three Fingers.}
    \label{tab:prediction performance table}
    \begin{tabular}{cclll}
\hline
                                         &                                    & Finger 1 & Finger 2 & Finger 3 \\ \hline
\multicolumn{1}{c|}{\multirow{2}{*}{$ F_{n} $}} & \multicolumn{1}{c|}{RMSE}          & 0.5660    & 1.3923   & 0.8525   \\
\multicolumn{1}{c|}{}                    & \multicolumn{1}{c|}{R$^2$}            & 0.9838   & 0.8800     & 0.9310   \\ \hline
\multicolumn{1}{c|}{\multirow{3}{*}{$ T_{z} $}} & \multicolumn{1}{c|}{RMSE}          & 0.0343   & 0.0547   & 0.0602   \\
\multicolumn{1}{c|}{}                    & \multicolumn{1}{c|}{R$^2$}            & 0.8739   & 0.8347   & 0.8147   \\
\multicolumn{1}{c|}{}                    & \multicolumn{1}{c|}{Success Rate} & 0.9844   & 0.9437   & 0.9725       \\ \hline
\end{tabular}

\end{table}

    As the results shown, the accuracy of the three fingers is different from each other, which is mainly due to the error caused by the limitation of finger making process. The accuracy indicators of each finger individually, RMSE and R$^2$, of $F_n$ is within acceptable limits for all three fingers. Though RMSE and R$^2$ of $T_z$ is not good, the success rate of the classification for $T_z$ which our focus is on shows great performance for all three fingers. Therefore, the calibration of the fingers is finished and the model is able to be used in further interactive grasping.

\begin{figure}[t]
    \begin{centering}
    \textsf{\includegraphics[width=0.5\columnwidth]{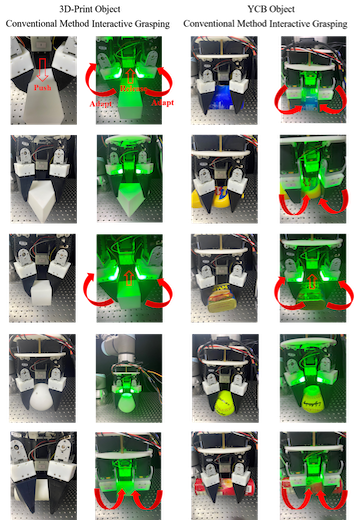} }
    \par\end{centering}
    \caption{Snapshots of the active adaptation to different types of objects.}
    \label{fig:result1}
\end{figure}

\subsection{Rigid-Soft Interactive Grasping}
    
    To test the proposed grasp policy's performance, grasping tests with the set of standard objects is conducted. Fig. \ref{fig:result1} shows snapshots of the active adaptation to different types of objects. The gripper transition from the circular configuration to parallel configuration when grasping cylinder object and to Lateral Configuration when grasping cube or cuboid object. For items with an irregular shape, an interactive grasping policy can conform to objects' shape and compute each finger's normal force for a robust grasp. 

    We try to grasp the target object one by one, placed on the table in a fixed position with a little randomness, five times for each object. After grasp completed, we shake the robot arm, inspired by \cite{pinto2017supervision}, to estimate this grasp's anti-disturbance ability. If the object is shaken or released on the table, this grasp is estimated a fail grasp. A conventional grasp policy that is grasping with only circular configuration is conducted for contrast. Furthermore, a set of YCB objects \cite{calli2015ycb} shown in Fig.\ref{fig:Overall} (d) are used to test the generalization of our grasp policy. Results are shown in Tab. \ref{tab:grasping result table}. In general, interactive grasping performs better than the conventional method, especially for the cuboid, cube, or other square objects. As for circular objects, the conventional method performs as better as interactive grasping.  

\begin{table}[]
	\centering
	\caption{Comparison of Success Rate by Conventional Method and Interactive Grasping}
	\label{tab:grasping result table}
	\begin{tabular}{l|lll}
	\hline
	                & Conventional Method & Interactive Grasping & Success Rate (\%) \\ \hline
	Cuboid          & 2/5                 & 5/5                  & 40/100            \\
	Triangular Prim & 5/5                 & 5/5                  & 100/100           \\
	Cube            & 1/5                 & 5/5                  & 20/100            \\
	Ball            & 5/5                 & 5/5                  & 100/100           \\
	Cylinder        & 5/5                 & 5/5                  & 100/100           \\
	Coffee Can      & 3/5                 & 4/5                  & 60/80             \\
	Mustered Bottle & 3/5                 & 4/5                  & 60/80             \\
	Potted Meat Can & 0/5                 & 5/5                  & 0/100             \\
	Tennis Ball     & 5/5                 & 5/5                  & 100/100           \\
	Chips Can       & 4/5                 & 5/5                  & 80/100            \\ \hline
	\end{tabular}
\end{table}

\section{Discussion}
\label{sec:Discussion}
\subsection{Calibration}
    
    Three fingers show an obvious different performance in the calibration process due to the fabrication error, which might cause a computation error for further grasp policy optimization. Also, the regression error of $F_n$ and $T_z$ is not negligible. The possible reason is that the dimension of input, in this paper is five, is not big enough for a high-resolution force sensor. However, humans can feel an object and show dexterous manipulation skills with their surface, not a high-resolution sensor. In this process, the sign of force or torque is more important rather than magnitude. Only a rough estimation of the sensor is enough. Inspired by this, we apply our sensor in rigid-soft interactive grasping and get better performance than the conventional method. Our grasping policy shows that optic fiber has great potential for sensing the deformation of the soft finger. 

    In the future, optoelectronic innervated soft gripper could be further explored to improve the prediction performance. Furthermore, Fiber Bragg Grating (FBG) is another possible choice for sensing soft fingers' deformation. Much research has demonstrated the ability of 3-dimensional reconstruction, which could be used in a soft finger. In addition to force or physical deformation, other information like temperature should also be considered. 

\subsection{Rigid-Soft Interactive Grasping}

    We apply our soft finger sensor in our rigid-soft interactive grasping process. During the execution, our finger can transmit base configuration between circular configuration, a parallel configuration, and lateral configuration to adapt the shape of target objects, shown in Fig. \ref{fig:result1}. The adaption brings a more robust grasp, which is proved in Table \ref{tab:grasping result table}, especially when grasping objects, which prefer a configuration transmission. In other words, in this paper, our initial grasp configuration is circular such that our grasp policy did not shows a noticeable distinction when dealing with circular objects versus the conventional method. Generally, there is no adjustment after the initial grasp in this process. 

    While as for cuboid objects, a parallel configuration is preferred. Although circular configuration can grasp such an object, this grasp has low anti-disturbance ability such that the object is likely to be released when moving quickly in high acceleration. In this grasp, the soft finger attaches the target object with a lateral pillar, manipulating the object.

    As for cubic objects, the lateral configuration is preferred. When using circular configuration to deal with such objects, the requirement of object position is strict. Once there is a random error in position estimation, a form of closure can not be achieved, and the object is pushed by one of these three fingers. As a result, the grasp is not stable and easy to be destroyed. Interactive grasping can compute the average force balance after adapting the surface of objects. One of three fingers should be released, and the other two fingers could achieve a force balance. Obviously, this grasp is more robust than before, and this is also shown in Table \ref{tab:grasping result table}.

\section{Conclusion}
\label{sec:Conclusion}

    In conclusion, this paper presents an optoelectronic innervated gripper for sensing the soft finger's deformation and interpreting this to force and torque information. Then we adapt this gripper for a grasp optimization policy. Based on a real-time force and torque estimation, though the estimation is rough, we get a better and more robust grasp that could resist interference. During the optimization progress, the gripper's base configuration is adjusted after the initial grasp and magnitude of force are adjusted for balance. 

    This work is a preliminary exploration of soft finger grasp optimization after initial grasping. In future research, we would like to investigate further our soft finger sensor's design for better performance, a more in-depth discussion of the optimization progress, and further exploration of multi-sensory information.

\bibliographystyle{unsrt}
\bibliography{references}  
\end{document}